\documentclass[runningheads,a4paper]{llncs}

\usepackage{pgfplots}
\usepackage{times}
\usepackage{graphicx}
\usepackage{epsfig}
\usepackage{xcolor}
\usepackage{url}

\urldef{\mailsa}\path|{llenc,pkral}@kiv.zcu.cz|

\begin{document}

\title {Deep Neural Networks for Czech Multi-label Document Classification}


\author{Ladislav Lenc\inst{1, 2} \and Pavel Kr\'al\inst{1, 2}}




\institute{
        Dept. of Computer Science \& Engineering\\
        Faculty of Applied Sciences\\
        University of West Bohemia\\
        Plze\v{n}, Czech Republic\\
         \and
        NTIS - New Technologies for the Information Society\\
        Faculty of Applied Sciences\\
        University of West Bohemia\\
        Plze\v{n}, Czech Republic\\
\mailsa\\
}

%
%


\toctitle{} \tocauthor{}
\maketitle

\hskip 150pt

\begin{abstract}
This paper is focused on automatic multi-label document classification of Czech text documents.
The current approaches usually use some pre-processing which can have negative impact (loss of information, additional implementation work, etc).
Therefore, we would like to omit it and use deep neural networks that learn from simple features. 
This choice was motivated by their successful usage in many other machine learning fields.
Two different networks are compared: the first one is a~standard multi-layer perceptron, while the second one is a~popular convolutional network.
The experiments on a~Czech newspaper corpus show that both networks significantly outperform baseline method which uses a rich set of features with maximum entropy classifier. 
We have also shown that convolutional network gives the best results.

\keywords{Czech, Deep Neural Networks, Document Classification, Multi-label}
\end{abstract}

\setcounter{footnote}{0}

\section{Introduction}
The amount of electronic text documents is growing extremely rapidly and therefore automatic document classification (or categorization) becomes very important for information organization, storage and retrieval.
Multi-label classification is considerably more important than the single-label classification because it usually corresponds better to the needs of the current applications.

The modern approaches usually use several pre-processing tasks: feature selection/reduction~\cite{compar_study}; precise document representation (e.g. POS-filtering, particular lexical and syntactic features, lemmatization, etc.)~\cite{multipleaa} to reduce the feature space with minimal negative impact on classification accuracy.
However, this pre-processing has several drawbacks as for instance loss of information, significant additional implementation work, dependency on the task/application, etc. 

Neural networks with deep learning are today very popular in machine learning field and
it was proved that they outperform many state-of-the-art approaches without any parametrization.
This fact is particularly evident in image processing~\cite{krizhevsky2012imagenet}, 
however it was further showed that they are also superior in Natural Language Processing (NLP) including Part-Of-Speech (POS) tagging, chunking, named entity recognition or semantic role labelling~\cite{collobert2011natural}.
However, to the best of our knowledge, the current published work does not include their application for multi-label document classification.

Therefore, the main goal of this paper consists in using neural networks for multi-label document classification of Czech text documents.
We will compare several topologies with different number of parameters to show that they can have better accuracy than the state-of-the-art methods.

We use and compare standard feed-forward networks (i.e. multi-layer perceptron) and popular Convolutional Networks (CNNs). 
To the best of our knowledge, this comparison was never been done on this task before.
Therefore, it is another contribution of this paper.
Note that we expect better performance of the CNNs as shown for instance in the OCR task~\cite{peyrard2015comparison}.



The results of this work should be integrated into an experimental multi-label document classification system.
The system should be used to replace manual annotation of the newspaper documents which is very expensive and time consuming task and thus save the human resources in the Czech News Agency (\v{C}TK)\footnote{http://www.ctk.eu}.


The rest of the paper is organized as follows.
Section~\ref{sec:RelWork}  is a short review of document classification methods with a~particular focus on neural networks. 
Section~\ref{sec:method} describes our document classification approaches. 
Section~\ref{sec:exps} deals with experiments realized on the \v{C}TK corpus and then discusses the obtained results.
In the last section, we conclude the experimental results and propose some future research directions.

\section{Related Work}
\label{sec:RelWork}

Document classification is usually based on supervised machine learning methods that exploit an annotated 
corpus to train a~classifier which then assigns the classes to unlabelled documents.
The most of works use Vector Space Model (VSM), which
usually represents each document with a~vector of all word occurrences weighted by their Term Frequency-Inverse Document Frequency (TF-IDF).

Several classification methods have been successfully used~\cite{Della}, for instance Bayesian classifiers, Maximum Entropy (ME), Support Vector Machines (SVMs), etc.
However, the main issue of this task is that the feature space in the VSM is highly dimensional which decreases the accuracy of the classifier.

Numerous feature selection/reduction approaches have been introduced~\cite{compar_study,lamirel2014optimizing} to solve this problem.
Furthermore, a~better document representation should help to decrease the feature vector dimension, 
e.g. using lexical and syntactic features as shown in~\cite{multipleaa}.
Chandrasekar et al. further show in~\cite{chandrasekar1996using} that it is beneficial to use POS-tag filtration in order to represent a~document more accurately.

More recently, some interesting approaches based on Latent Dirichlet Allocation (L-LDA)~\cite{Ramage2009} have been introduced.
Another method exploits partial labels to discover latent topics~\cite{Ramage:2011}.
Principal Component Analysis~\cite{Gomez12} incorporating semantic concepts~\cite{Yun12} has also  been used for the document classification. 

Recently, ``deep'' Neural Nets (NN) have shown their superior performance in many natural language processing tasks including POS tagging, chunking, named entity recognition and semantic role labelling~\cite{collobert2011natural} without any parametrization. 
Several different topologies and learning algorithms were proposed.

For instance, the authors of~\cite{zhang2015text} propose two Convolutional Neural Nets (CNN) for ontology classification, sentiment analysis and single-label document classification. 
Their networks are composed of 9 layers out of which 6 are convolutional layers and 3
fully-connected layers with different numbers of hidden units and frame sizes.
They show that the proposed method significantly outperforms the baseline approaches (bag of words) on English and Chinese corpora.
Another interesting work~\cite{kim2014convolutional} uses in the first layer (i.e. lookup table) pre-trained vectors from word2vec~\cite{mikolov2013efficient}. 
The authors show that the proposed models outperform the state-of-the-art on 4 out of 7 tasks, which
include sentiment analysis and question classification.
 

For additional information about architectures, algorithms, and applications of deep learning, please refer the survey~\cite{Deng2014}.

On the other hand, classical feed-forward neural nets architectures represented particularly by multi-layer perceptrons are used rather rarely.
However, these models were very popular before and some approaches for document classification exist.
Manevitz et al. show in~\cite{Manevitz20071466} that their simple feed-forward neural network with three layers (20 inputs, 6 neurons in hidden layer and 10 neurons in the output layer, i.e. number of classes) gives F-measure about 78\% on the standard Reuters dataset.


Traditional multi-layer neural networks were also used for multi-label document classification in~\cite{zhang2006multilabel}.
The authors have modified standard backpropagation algorithm for multi-label learning which employs a novel error function.
This approach is evaluated on functional genomics and text categorization.

The most of the proposed approaches is focused on English and only few works deal with Czech language.
Hrala et al. use in~\cite{hrala13} lemmatization and Part-Of-Speech (POS) filtering for a~precise representation of Czech documents.
In~\cite{Kral13}, three different multi-label classification approaches are compared and evaluated.
Another recent work proposes novel features based on the unsupervised machine learning~\cite{Kral14MICAI}.
To the best of our knowledge, no document classification approach using neural nets deals with Czech language.

\section{Neural Nets for Multi-label Document Classification}
\label{sec:method}

\subsection{Baseline Classification}
The feature set is created according to Brychc\'in et al.~\cite{Kral14MICAI} and is composed of words, stems and features created by S-LDA and COALS.
They are used because the authors experimentally proved that the additional unsupervised features significantly improve classification results.

For multi-label classification, we use an efficient approach presented by Tsoumakas et al. in~\cite{tsoumakas2007multi}.
This method employs $n$ binary classifiers $C_{i=1}^n: d \rightarrow {l,\neg l}$ 
(i.e. each binary classifier assigns the document $d$ to the label $l$ 
iff the label is included in the document, $\neg l$ otherwise).
The classification result is given by the following equation:

\begin{equation}
C(d)=\cup_{i=1}^n{:C_i(d)}
\end{equation}

The Maximum Entropy (ME) model is used for classification.

\subsection{Standard Feed-forward Deep Neural Network (FDNN)}
Feed-forward neural networks are probably the most commonly used type of NNs.
We propose to use an MLP with two hidden layers which can be seen as a deep network\footnote{We have also experimented with an MLP with one hidden layer with lower accuracy.}. 
As an input of our network we use the simple Bag of Words (BoW) which is
a~binary vector where value 1 means that the word with a given index is present in the document. 
The size of this vector depends on the size of the dictionary which is limited by $N$ most frequent words.
The only preprocessing is the conversion of all characters to lower case and also replacing of all numbers by one common token.
 
The size of the input layer thus depends on the size of the dictionary that is used for the feature vector creation.
The first hidden layer has 1024 while the second one has 512 nodes\footnote{This configuration was set experimentally.}.
The output layer has size equal to the number of categories which is 37 in our case.
To handle the multi-label classification, we threshold the values of nodes in the output layer. 
Only the values larger than a given threshold are assigned to the labels.

\subsection{Convolutional Neural Network (CNN)}
The input feature of the CNN is a~sequence of words in the document.
We use similar document preprocessing and also similar dictionary as in the previous approach.
The words are then represented by the indexes into the dictionary.

The first important issue of this network for document classification is variable length of documents.
It is usually solved by setting a~fixed value and longer documents are shortened while shorter ones must be padded to ensure exactly the same length.
The words that are not in the dictionary are assigned to a~reserved index and the padding has also a~reserved index.

The architecture of our network is motivated by Kim in~\cite{kim2014convolutional}.
However, we use just one size of the convolutional kernel and not the combination of several sizes. 
Our kernels have only 1 dimension (1D) while Kim have used larger 2 dimensional kernels.
This is mainly due to our preliminary experiments where the simple 1 dimensional kernels gave better results than the larger ones. 

The input of our network is a vector of word indexes of the length $L$ where $L$ is the number of words used for document representation. 
The second layer is an embedding layer which represents each input word as a~vector of a given length.
The document is thus represented as a~matrix with $L$ rows and $EMB$ columns where $EMB$ is the length of embedding vectors.
The third layer is the convolutional one. 
We use $ N_C $ convolution kernels of the size $K \times 1$ which means we do 1D convolution over one position in the embedding vector over $K$ input words. 
The following layer performs max pooling over the length $ L - K + 1 $ resulting in $ N_C $ \ $ 1 \times EMB $ vectors. 
The output of this layer is then flattened and connected with the output layer containing 37 nodes. 

\begin{figure}[!htb]
  \centering
  {\epsfig{file=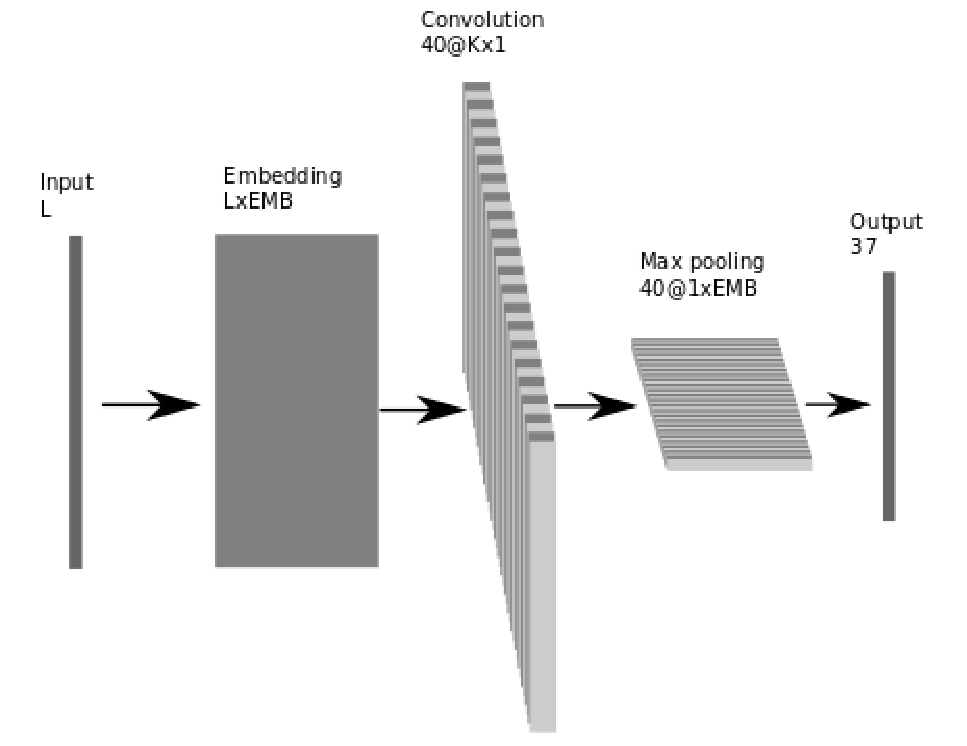,width=8cm,angle=0}}
   \caption{Architecture of the convolutional network}
  \label{fig:cnn}
\end{figure}

The output of the network is then thresholded to get the final results.
The values greater than a~given threshold indicate the labels that are assigned to the classified document. 
The architecture of the network is depicted in Figure~\ref{fig:cnn}.

\section{Experiments}
\label{sec:exps}
In this section we first describe the Czech document corpus that we used for evaluation of our methods. 
After that we describe the performed experiments and the final results.
The results are compared with previously published results on the Czech document corpus. 

\subsection{Tools and Corpus}
\label{sec:tools}
For implementation of all neural-nets we used Keras tool-kit~\cite{chollet2015} which is based on the Theano deep learning library~\cite{bergstra2010theano}.
It has been chosen mainly because of good performance and our previous experience with this tool.
All experiments were computed on GPU to achieve reasonable computation times.

\begin{figure}[!htb]
  \centering
  {\epsfig{file=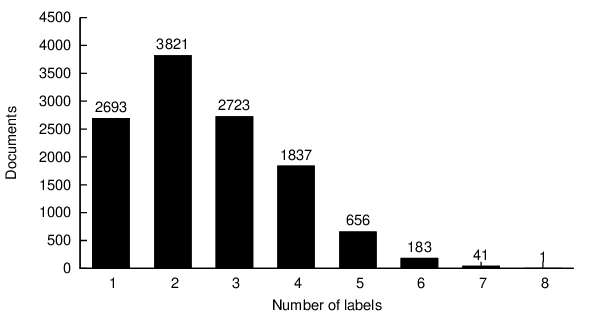,width=8cm,angle=0}}
   \caption{Distribution of documents depending on the number of labels}
  \label{fig:doc_labels}
\end{figure}

\begin{figure}[!htb]
  \centering
  {\epsfig{file=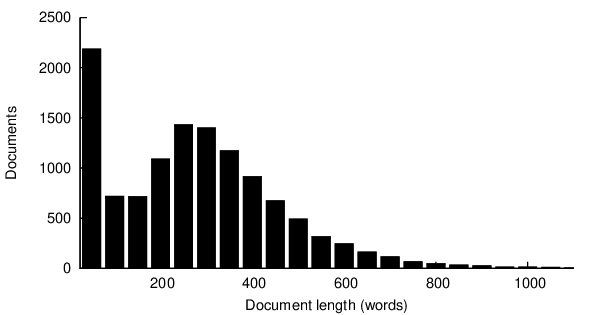,width=10cm,angle=0}}
   \caption{Distribution of the document lengths}
  \label{fig:doc_lengths}
\end{figure}

As already stated, the results of this work shall be used by the \v{C}TK.
Therefore, for the following experiments we used the Czech text documents provided by the \v{C}TK.
This corpus contains 2,974,040 words belonging to 11,955 documents.
The documents are annotated from a~set of 60 categories out of which we used 37 most frequent ones.
The category reduction was done to allow comparison with previously reported results on this corpus where the same set of 37 categories was used. 
Figure~\ref{fig:doc_labels} illustrates the distribution of the documents depending on the number of labels.
Figure~\ref{fig:doc_lengths} shows the distribution of the document lengths (in word tokens).
This corpus is freely available for research purposes at~\url{http://home.zcu.cz/~pkral/sw/}. 

We use the five-folds cross validation procedure for all following experiments,
where 20\% of the corpus is reserved for testing and the remaining part for training of our models. 
For evaluation of the document classification accuracy, we use the standard 
F-measure ({\it F1}) metric~\cite{powers2011evaluation}. 
The confidence interval of the experimental results is~0.6\% at a~confidence level of 0.95~\cite{press1996numerical}.

\subsection{Experimental Results}

\subsubsection{FDNN}
\begin{figure}
  \centering
  {\epsfig{file=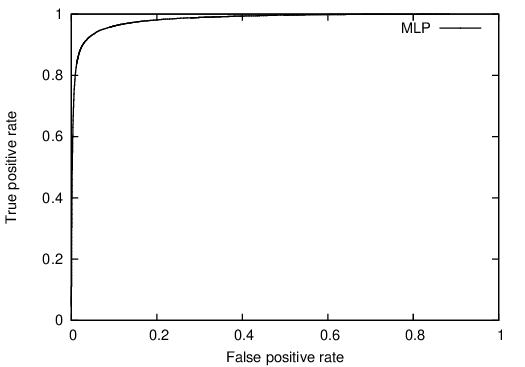,width=8cm,angle=0}}
   \caption{ROC curve of the FDNN}
  \label{fig:roc_fdnn}
\end{figure}

As a~first experiment, we would like to validate the proposition of thresholding applied to the output layer of the FDNN. 
For this task we use the Receiver Operating Characteristic (ROC) curve which clearly shows the relationship between the true positive and the false positive rate for different values of the {\it acceptance} threshold.
We use 20,000 most common words to create the dictionary.
The ROC curve is depicted in Figure~\ref{fig:roc_fdnn}.
According to the shape of this curve we can conclude that the proposed approach is suitable for multi-label document classification.

In the second experiment we would like to identify the optimal activation function of the nodes in the output layer.
Two functions (sigmoid and softmax) are compared and evaluated. 
We have evaluated the threshold values in interval $[0;1]$, 
however only the best classification scores are depicted (see Table~\ref{tab:activation_fdnn}, best threshold values in brackets).
This table shows that the softmax gives better results. 
Based on these results, we will further use this activation function and the threshold is set to 0.11.


\begin{table}[h]
\caption{Comparison of output layer activation functions of the FDNN (threshold values depicted in brackets)}
\label{tab:activation_fdnn}
\centering
\begin{tabular}{c|c}
\textbf{Activation function} & \textbf{F1 [\%]} \\
\hline
\hline softmax & 83.8 (0.11) \\
\hline sigmoid & 82.3 (0.48) \\
\end{tabular}
\end{table}

The third experiment studies the influence of the dictionary size on the performance of the FDNN. 
Table~\ref{tab:fdnn_words} shows the dependency of F-measure on the word number in the dictionary.
This table shows that the previously chosen 20,000 words is a~reasonable choice and further increasing  the number does not bring any {\it significant} improvement. 

\begin{table}[h]
\caption{F-measure of FDNN with different numbers of words in the dictionary}
\label{tab:fdnn_words}
\centering
\begin{tabular}{c||c|c|c|c|c|c|c|c}
\textbf{Word number} & \ 1,000 \ & \ 2,000 \ & \ 5,000 \ & \ 10,000 \ & \ 15,000 \ & \ 20,000 \ & \ 25,000 \ & \ 30,000 \ \\
\hline \textbf{F1 [\%]} & 72.1 & 76.7 & 80.9 &  82.9 &  83.5 & 83.8 & 83.9 & 83.9 \\
\end{tabular}
\end{table}

\subsubsection{CNN}
\begin{figure}
  \centering
  {\epsfig{file=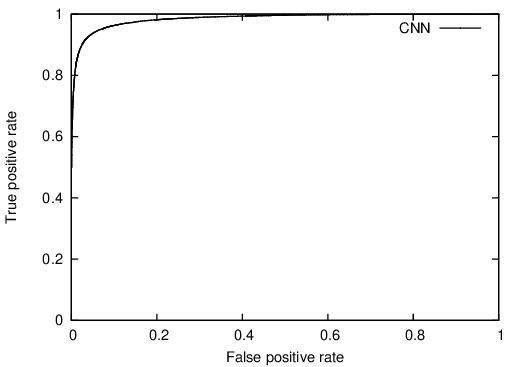,width=8cm,angle=0}}
   \caption{ROC curve of the CNN}
  \label{fig:roc_cnn}
\end{figure}

In all experiments performed with the CNN we use the same dictionary size (20,000 words) as in the case of FDNN to allow a straightforward comparison of the results.
According to the analysis of our corpus we estimate that a~suitable vector size for document representation is 400 words.
As well as for the FDNN we first compute the ROC curve to validate the proposition of thresholding in the output.
Figure~\ref{fig:roc_cnn} clearly shows that this approach is suitable for our task.

As a~second experiment we identify an optimal activation function of neurons in the output layer. 
We compare the softmax and sigmoid functions.
The achieved F-measures are depicted in Table~\ref{tab:cnn_activation}.
It is clearly visible that in this case the sigmoid function performs better.
We will thus use the sigmoid activation function and the threshold will be set to 0.1 for all further experiments with CNN. 

\begin{table}[h]
\caption{Comparison of output layer activation functions of the CNN (threshold values depicted in brackets)}
\label{tab:cnn_activation}
\centering
\begin{tabular}{c|c}
\textbf{Activation function} & \textbf{F1 [\%]} \\
\hline
\hline softmax & 80.9 (0.07) \\
\hline sigmoid & 84.0 (0.10) \\
\hline 
\end{tabular}
\end{table}

In this experiment, we will show the impact of the number of convolutional kernels in our network on the classification score.
400 words are used for document representation ($L = 400$) and the embedding vector size is 200.
This experiment shows (see Table~\ref{tab:kernel_number}) that this parameter influences the classification score only very slightly ($\Delta F1 \sim +1\%$).
All values from interval [20; 64] are suitable for our goal 
and therefore we chose the value of 40 for further experimentation.

\begin{table}[h]
\caption{F-measure of CNN with different numbers of convolutional kernels}
\label{tab:kernel_number}
\centering
\begin{tabular}{c||c|c|c|c|c|c|c|c|c|c|c|c|c|c}
\textbf{Kernel no.} & \ \ 12 \ \ &  \ \ 16 \ \ & \ \ 20 \ \ & \ \ 24 \ \ & \ \ 28 \ \ & \ \ 32 \ \ & \ \  36 \ \ & \  \ 40  \ \ &  \ \ 44 \ \  & \ \  48 \ \ &  \ \ 52 \ \ &  \ \ 56 \ \  & \ \ 60 \ \ &  \ \ 64  \ \ \\
\hline \textbf{F1 [\%]} & 83.1 & 83.4 & 84.0 & 83.9 & 84.1 & 84.1 & 84.1 & 84.1 & 84.1 & 84.2 & 84.2 & 84.1 & 84.1 & 84.0\\
\end{tabular}
\end{table}

\begin{figure}
  \centering
  {\epsfig{file=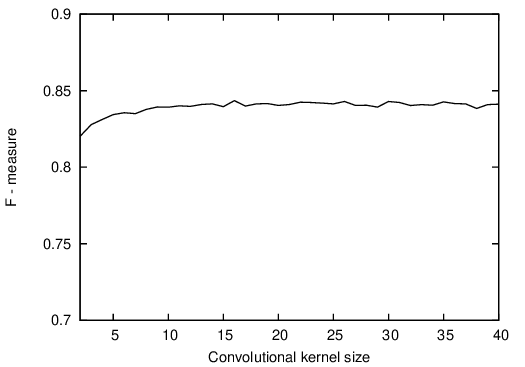,width=8cm,angle=0}}
   \caption{Dependency of F-measure on the size of convolutional kernel}
  \label{fig:cnn_kernel}
\end{figure}

The following experiment shows the dependency of F-measure on the size of convolutional kernels.
We use 40 kernels and
the size of the kernel varies from 2 to 40. 
The size of the kernels can be interpreted as the length of word sequences that the CNN works with.
Figure~\ref{fig:cnn_kernel} shows that the results are comparable and as a~good compromise we chose the size of 16 for the following experiments.

Finally, we tested our network with different numbers of input words and with varying size of the embedding vectors.
Table~\ref{tab:cnn} shows the achieved results with several combinations of these parameters.
We can conclude that the 400 words that we chose at the beginning was a~reasonable choice. 
However, it is beneficial to use longer embedding vectors. 
It must be noted that the further increasing of the embedding size has a strong impact on the computation time and might be not practical for real-world applications.

\begin{table}[h]
\caption{F-measure of CNN with different word numbers and different embedding sizes [\%]}
\label{tab:cnn}
\centering
\begin{tabular}{c||c|c|c|c|c}
\textbf{Word number} & \ \ \ 100 \ \ \ & \ \ \ 200 \ \ \ & \ \ \ 300 \ \ \ & \ \ \ 400 \ \ \ & \ \ \ 500 \ \ \ \\
\textbf{Embedding length} & & & & & \\
\hline
\hline 100 & 82.3 & 82.7 & 83.1 & 83.3 & 83.4 \\
\hline 150 & 82.9 & 83.3 & 83.4 & 83.7 & 83.9 \\
\hline 200 & 82.9 & 83.4 & 83.8 & 84.0 & 84.1 \\
\hline 250 & 82.5 & 83.8 & 84.1 & 84.2 & 84.4 \\
\hline 300 & 83.2 & 83.9 & 84.3 & 84.3 & 84.5 \\
\hline 350 & 83.4 & 83.9 & 84.3 & 84.6 & 84.5 \\
\hline 400 & 83.4 & 83.9 & 84.4 & 84.7 & 84.7 \\
\hline 450 & 83.4 & 83.9 & 84.5 & 84.8 & 84.3 \\
\end{tabular}
\end{table}


\subsubsection{Summary of the Results}
Table~\ref{tab:final} compares the results of our approaches with another efficient method~\cite{Kral14MICAI}.
The results show that both proposed approaches significantly outperform this baseline approach that uses several features with ME classifier.

\begin{table}[h]
\caption{Comparison of the results of our approaches with maximum entropy based method}
\label{tab:final}
\centering
\begin{tabular}{c|c|c|c}
\textbf{Method} & \textbf{Precision} & \textbf{Recall} & \textbf{F1 [\%]} \\
\hline
\hline Brychc\'in et al.~\cite{Kral14MICAI} & 89.0 & 75.6 & 81.7 \\
\hline
\hline FDNN & 83.7 & 83.6 & 83.9 \\
\hline CNN & 86.4 & 82.8 & 84.7 \\ 
\end{tabular}
\end{table}


\section{Conclusions and Future Work}
In this paper, we have used two different neural nets for multi-label document classification of Czech text documents. 
Several experiments were realized to set optimal network topologies and parameters.
An important contribution is the evaluation of the performance of neural networks using simple features. 
Therefore we have used the BoW representation for the FDNN and sequence of word indexes for the CNN as the inputs.
Based on these experiments we can conclude:

\begin{itemize}
\item the two proposed network topologies together with thresholding of the output are efficient for multi-label classification task
\item softmax activation function is better for FDNN, while sigmoid activation function gives better results for CNN
\item CNN outperforms FDNN only very slightly ($\Delta$ F1 $\sim +0.6\%$)
\vskip 5pt
\item the most important is the fact that both neural nets with only basic pre-processing and without any parametrization significantly improve the baseline maximum entropy method with a rich set of parameters ($\Delta$ F1 $\sim +4\%$)
\end{itemize}

Based on these results, we want to integrate CNN into our experimental document classification system.

In this paper, we have used relatively simple convolution neural network. 
Therefore, our first perspective consists in designing a more complicated CNN architecture.
According to the literature, we assume that more layers in this network will have a positive impact on the classification score.
Our embedding layer was also not initialized by some pre-trained semantic vectors (e.g. word2vec or GloVe).
Another perspective thus consists in initializing of the embedding CNN layer with pre-trained vectors.


\section*{Acknowledgements}
This work has been supported by the project LO1506 of the Czech Ministry of Education, Youth and Sports.
We also would like to thank Czech New Agency (\v{C}TK) for support and for providing the data.

\bibliographystyle{splncs}

\bibliography{biblio,multi-label,doc_ner,ner,mlsp14}

\end{document}